\documentclass[letterpaper, 10 pt, conference]{ieeeconf}  
\usepackage[utf8]{inputenc}
\usepackage[natbib=true,style=numeric,sorting=nyt]{biblatex}
\usepackage{algorithm}
\usepackage{algpseudocode}
\usepackage{caption}
\usepackage{subcaption}
\usepackage{graphicx}
\usepackage{multirow}
\usepackage{pgfplots}
\usepackage{pgfplotstable}
\usepackage{booktabs}
\usepackage{amsmath}
\pgfplotsset{compat=1.16}
\usepackage{acronym}
\usepackage{etoolbox}
\usepackage{hyperref}
\usepackage{fancyhdr}

\usepackage{amsmath}
\usepackage{booktabs}
\usepackage{hyperref}

\title{\LARGE \bf
Zwitscherkasten - DIY Audiovisual bird monitoring\\
\vspace{1mm}
\small\url{https://github.com/cvims/zwitscherkasten}
}

\author{
\authorblockN{
Dominik Blum$^{*,1}$, 
Elias Häring$^{*,1}$, 
Fabian Jirges$^{*,1}$, 
Martin Schäffer$^{*,1}$, 
David Schick$^{*,1}$, \\
Florian Schulenberg$^{*,1}$, 
Torsten Schön$^{1}$
}
\authorblockA{
$^{1}$Technische Hochschule Ingolstadt, Ingolstadt, Germany \\
$^{*}$Equal contribution.
}
}

\begin{document}
\pagestyle{plain}      

\acrodef{ML}{Machine Learning}
\acrodef{CV}{Computer Vision}


\maketitle

\begin{abstract}

This paper presents Zwitscherkasten, a DiY, multimodal system for bird species monitoring using audio and visual data on edge devices. Deep learning models for bioacoustic and image-based classification are deployed on resource-constrained hardware, enabling real-time, non-invasive monitoring. An acoustic activity detector reduces energy consumption, while visual recognition is performed using fine-grained detection and classification pipelines. Results show that accurate bird species identification is feasible on embedded platforms, supporting scalable biodiversity monitoring and citizen science applications.

\end{abstract}


\section{Introduction}

Avian populations in Europe, particularly in Germany, are experiencing severe declines driven by habitat fragmentation and anthropogenic pressures. Over one-third of German bird species are classified as endangered, highlighting the need for scalable, non-invasive monitoring approaches~\cite{Hoechst2022}. Passive monitoring enables large-scale biodiversity assessment, but traditional cloud-based pipelines suffer from latency, bandwidth, privacy, and connectivity limitations. Edge computing addresses these issues by enabling real-time, low cost species identification directly on edge devices such as Raspberry Pi and Rubik Pi.

This work presents a multimodal avian monitoring system optimized for home deployment in german ecosystems meant to classify birds seen throughout Germany. We (1) develop and benchmark deep learning models for both visual and audio-based bird species classification, comparing CNN- and transformer-based audio classifiers alongside transfer-learned visual models tailored to German avifauna; (2) introduce an audio intent model that decouples bird presence detection from fine-grained species classification to improve robustness in real-world acoustic environments and decrease energy-consumption in long-time deployments; (3) incorporate interpretability methods to expose discriminative spectral-temporal features underlying audio classification decisions; and (4) deploy the resulting multimodal system on embedded hardware platforms, including Raspberry~Pi and Rubik~Pi, with an integrated human-machine-interface (HMI), analyzing accuracy through model sizes. The paper represents a framework for Do-It-Yourself (DiY) avian monitoring. Our results demonstrate that state-of-the-art multimodal bird classification is feasible on low-power hardware, enabling scalable biodiversity monitoring and citizen science applications.


\section{Related Work}
Bird species recognition has been extensively studied across photographs and audio recordings. Image-based approaches leverage fine-grained visual cues such as plumage, shape, and context, whereas bioacoustic methods exploit species-specific vocal signatures that are often robust when visual observations are limited (e.g., dense vegetation or low light). Although these research lines are frequently treated separately, they address the same underlying fine-grained classification problem and face similar challenges such as long-tailed data and inter-class similarity. In the following, we review prior work on photo-based bird classification and audio-based bioacoustic recognition to position our approach within the broader literature.

\subsection{Audio based bird classification}

The automated classification of bird vocalizations has seen rapid progress in recent years, driven by advances in deep learning architectures, large-scale curated datasets, and the growing availability of computational resources. This section reviews the current state-of-the-art approaches, focusing on the principal architectural families, key datasets, and performance benchmarks that contextualize modern bird audio classification systems.

\subsubsection{Convolutional Neural Network Approaches}

Convolutional neural networks have historically dominated bird audio classification tasks, leveraging their proven effectiveness on spectrogram data. BirdNET~\cite{kahl2021}, developed at Cornell Lab of Ornithology, represents a landmark achievement: a ResNet-based architecture with 157 layers and over 27 million parameters, trained to identify 984 North American and European bird species. The BirdNET pipeline processes raw acoustic signals into log-scaled Mel-spectrograms, segments audio into 3-second windows to capture temporal vocalization structure, and applies biogeographical contextual filtering using location and temporal metadata to refine predictions. On independent soundscape datasets, BirdNET achieved a mean average precision (mAP) of 0.791 for single-species recordings and an F$_{0.5}$ score of 0.414 for complex annotated soundscapes.

More recent CNN-based approaches have adopted efficient architectures optimized for downstream deployment. EfficientNet~\cite{tan2019}, a family of models that rethinks scaling for convolutional networks, has been successfully applied to bird sound classification, achieving accuracy levels competitive with or exceeding ResNet-50 (approximately 85-90\% on curated datasets).  For edge deployment, lightweight variants such as MobileNet and specialized systems like BirdNET-Pi enable real-time bird identification on resource-constrained devices (e.g., Raspberry Pi, Jetson Nano platforms).

\subsubsection{Transformer-Based and Foundation Model Approaches}

The application of transformer architectures to audio spectrograms represents a significant methodological shift in bird audio classification. The Patchout fast Spectrogram Transformer (PaSST)~\cite{koutini2022} addresses the computational bottleneck inherent in transformer-based audio processing: the quadratic scaling of attention complexity with sequence length. PaSST implements Patchout regularization, which selectively drops spectrogram patches during training, simultaneously reducing computational and memory requirements while improving robustness. Pre-trained on AudioSet, PaSST achieves competitive performance (mAP 47.1 on AudioSet-2M) and trains comparatively efficiently on single GPUs, enabling practical downstream fine-tuning \cite{passt}.

Self-supervised learning has emerged as a powerful paradigm for audio representation learning. HuBERT~\cite{hsu2021hubert} pioneered masked prediction objectives for speech, influencing subsequent bioacoustics models. BirdAVES, released by the Earth Species Project (2024), represents a self-supervised foundation model explicitly tailored for avian vocalizations \cite{hagiwara2022avesanimalvocalizationencoder}. Building on masked unit prediction, BirdAVES generates fine-grained, per-frame embeddings (50 ms resolution) without requiring explicit species labels. Remarkably, BirdAVES achieves over 20\% performance improvement on bird-focused benchmarks compared to its predecessor (AVES) and approaches or exceeds the performance of supervised models such as BirdNET and Perch in zero-shot and few-shot transfer scenarios despite using only a fraction of the available labeled data.

\subsubsection{Datasets and Benchmarks}

The rapid advancement of bird audio classification has been enabled by large-scale, curated, and publicly accessible datasets. Xeno-Canto (launched 2005) remains the world's largest citizen-science repository of wildlife sounds, containing over 1 million recordings spanning more than 12,900 species, each accompanied by spectrogram visualizations, geographic coordinates, and quality ratings \cite{vellinga2015}.

BirdSet~\cite{rauch2024} provides a comprehensive benchmark that substantially exceeds AudioSet in scale and bird-specific diversity. BirdSet comprises 6,800+ hours of training audio from nearly 10,000 species, derived primarily from Xeno-Canto's focal recordings. Critically, it includes eight strongly annotated evaluation datasets (448 total hours) across geographically and ecologically diverse regions (continental USA, Hawaii, Peru, Colombia), with multi-label classification tasks.

The Macaulay Library at Cornell Lab of Ornithology \cite{MacaulayLibrary}, established in 1929, provides one of the world's most comprehensive scientific archives of wildlife recordings, containing over 175,000 audio recordings covering 75\% of the world's bird species, complemented by over 50,000 video clips spanning more than 3,500 species. The library serves as both a critical training and validation resource for many contemporary models and provides curated reference guides (e.g., Cornell Guide to Bird Sounds with 4,800+ remastered audio files across 900 North American species).

The Tierstimmenarchiv (Animal Sound Archive) at Berlin's Museum für Naturkunde \cite{TierstimmenarchivMfN}, founded in 1951, represents a European resource, comprising approximately 120,000 bioacoustical recordings of 1,800 bird species, 580 mammal species, and numerous other fauna. One-third of recordings are field recordings, predominantly from Central and Eastern Europe, making it a complementary source to global datasets, particularly for training models on European avifauna with high-quality digitization (96 kHz, 24-bit resolution).

BEANS (Benchmark of Animal Sounds)~\cite{stowell2022} provides a complementary standardized benchmark comprising 12 datasets covering birds, terrestrial and marine mammals, anurans, and insects, with both classification and detection tasks. The Cornell Bird Identification dataset within BEANS spans 264 species with non-overlapping recordists between train and test splits, closely approximating real-world deployment scenarios.

\subsubsection{Feature Extraction and Augmentation}

Across all surveyed approaches, Mel-spectrograms constitute the dominant signal representation, offering an interpretable frequency-time decomposition that aligns with human auditory perception and avian vocalization acoustics. Models typically employ short-time Fourier transforms (STFT) with standard parameters (e.g., 1024-point FFT) to generate spectrograms, then convert to log-scale Mel-frequency bins (typically 64-128 channels), which are then used as input for classificiation of audio.

SpecAugment~\cite{park2019specaugment}, which applies random time-frequency masking to spectrograms, is a standard augmentation technique. Additional augmentations include time and frequency shifting, pitch shifting, and gain adjustments. Models pre-trained on general-purpose corpora (e.g., AudioSet) benefit from transfer learning, where pre-trained weights are fine-tuned on bird-specific data, often with frozen early layers to preserve learned feature hierarchies.

\subsubsection{Performance Synthesis and Emerging Trends}

Contemporary benchmarking efforts reveal a clear trajectory: foundation models and self-supervised approaches increasingly outperform purely supervised CNNs, especially in low-data regimes and on rare or geographically novel species. BirdCLEF competition results consistently show that ensemble approaches combining multiple CNN backbones (EfficientNet, ResNet) with transformer models yield superior results, albeit at increased computational cost.

Computational efficiency remains a practical concern. PaSST's Patchout regularization, lightweight CNN architectures, and edge-compatible systems (BirdNET-Pi, mobile variants) extend the applicability of bird audio classification to resource-constrained field deployments.

\subsection{Visual bird classification}

Visual bird species classification is an active and challenging research area within computer vision, sitting at the intersection of fine-grained visual categorization and object detection. 
Traditional image classification approaches treat the problem as a single-label prediction task. However, these methods often assume one object per image, which limits their utility in ecological datasets where multiple birds may co-occur within a scene. Fine-grained visual categorization research has established a large body of work on discriminating subordinate categories such as bird species using deep learning, including specialized architectures, attention mechanisms, and region-part learning \cite{wei2021fine, wang2023fine}.

Mainly, two constraints determined the considered approaches. 
\begin{itemize}
    \item Continuous monitoring of the scene using a camera must be ensured.
    \item Deployment on an edge device with limited CPU performance must be ensured (e.g. Raspberry Pi and Rubik Cube platforms).
\end{itemize}

Therefore, we decided to compare two different approaches.
\begin{itemize}
    \item Multiclass object detection models, where a single network simultaneously localizes and classifies instances.
    \item Two-stage pipelines, combining dedicated detectors with subsequent fine-grained classifiers.
\end{itemize}

\paragraph{Multiclass Object Detection}  
This approach aims for an end-to-end training framework. The problem of classifying a fine-grained bird species is reduced to an object detection problem. 
Single-stage object detectors such as the YOLO family (You Only Look Once) have become a de facto standard for joint localization and classification due to their real-time performance and end-to-end training frameworks \cite{redmon2016yolo, ali2024yolo}. Recent works demonstrate that YOLO-based models can be adapted for fine-grained bird detection and species identification in complex ecological scenes by exploiting enhanced feature extraction modules or domain-specific architectures. Furthermore, it characterizes the limits of a single-shot detector developed for objects that are well-separable  \cite{yi2023research}.

\paragraph{Two-Stage Detection–Classification Pipelines}  
An alternative strategy decouples detection and classification, where an object detector first localizes the bird, and a separate fine-grained classifier predicts the species. Félix-Jiménez et al. (2025) proposed a lightweight object detector (YOLOv8 Small) performs bird detection, and a separate MobileNet V3 classifier predicts species labels on the cropped detections. Their results demonstrate that such a decoupled pipeline achieves high detection and classification accuracy while remaining suitable for mobile and resource-limited environments. Due to the above-mentioned constraints, the object detector is additional used as an event detector \cite{felix2025yolov8mobilenet}. In ecological monitoring, such pipelines often leverage pretrained CNN backbones for classification, benefiting from transfer learning \cite{vanhorn2018inat}.

Conceptual distinctions in the literature suggest inherent trade-offs between these paradigms. 
End-to-end multiclass detection models offer simpler deployment and unified optimization, whereas two-stage detection–classification pipelines provide greater flexibility for fine-grained recognition by allowing specialized classifiers to operate on localized object regions, at the cost of increased system complexity and computational overhead \cite{wei2021fine}. 

Despite significant progress in visual bird recognition, many existing approaches are primarily evaluated on curated academic benchmarks such as CUB-200-2011 \cite{wah2011cub}, which offer controlled imaging conditions and limited ecological variability. 
Such datasets do not fully reflect the challenges of real-world bird observation scenarios.
This motivates the comprehensive evaluation presented in this study, which compares multiclass object detection and two-stage detection–classification strategies within a realistic wildlife monitoring setting using a large-scale, taxonomically curated dataset derived from iNaturalist observations.


\section{System Design and Implementation}

We propose a dual, parallel classification architecture comprising complementary audio and visual processing streams. For the acoustic modality, a lightweight CNN performs acoustic activity detection and serves as a gating mechanism, activating a more computationally intensive audio classifier only when avian vocalizations are detected. This design reduces unnecessary computation and energy consumption during periods of acoustic inactivity. In parallel, a visual classification network is employed to detect birds in close proximity to the camera sensor. The outputs of both modalities are fused at a late stage via the human–machine interface (HMI), which provides users with real-time classification results as well as historical summaries.

The system is deployed on resource-constrained embedded platforms, including the Rubik Pi and Raspberry Pi, demonstrating a feasible and energy-efficient solution for DiY avian monitoring. Furthermore, we show that the proposed models can be adapted for deployment on mobile devices, such as the Apple iPhone, highlighting the portability and broader applicability of the approach.

\subsection{Audio bird classification}

\subsubsection{Data Selection and Preprocessing}

The efficacy of deep learning in bioacoustics is heavily predicated on the quality and representation of the input features. We select \textbf{Xeno-canto} (XC) \cite{vellinga2015} as our primary data source, due to its open API access, coverage of german birds and availability of raw audio. To ensure both taxonomic coverage and model robustness, we select recordings from quality categories `A', `B', and `C'. While categories `A' and `B' provide high-fidelity signals, the inclusion of category `C'-characterized by lower signal-to-noise ratios and diverse background interference-was intentional. This diversity exposes the model to a wider manifold of acoustic conditions, representing realistic home-scenarios and preventing the model from over-fitting to idealized studio-quality samples.

Beyond XC, we evaluated several alternative repositories to assess their suitability for our classification task, particularly regarding Central European species coverage and availability:

\begin{itemize}
    \item \textbf{Macaulay Library:} As one of the world's largest archives of animal sounds, the Macaulay Library offers high-quality curated specimens. While its global reach is unparalleled, XC was favored for its open-access API structure and extensive community-driven labeling for rare European variants.
    
    \item \textbf{Tierstimmen-Archiv (TSA):} Based at the Museum für Naturkunde Berlin, the TSA \cite{frommolt2012} provides great coverage of German avian fauna. However, a significant portion of this archive utilizes analog-era recording practices where human voice-over commentary is embedded directly within the audio tracks. This presence of overlapping human speech presents a substantial obstacle for automated segmentation for training-data generation, making it less conducive to high-throughput deep learning pipelines compared to the metadata-isolated recordings of XC.
    
    \item \textbf{BirdSet:} We considered the \textit{BirdSet} benchmark \cite{rauch2024}, which serves as a large-scale, standardized framework for deep learning. It excels in providing a unified structure for multi-label classification, yet its fixed nature is less flexible for species-specific expansion compared to direct repository scraping.
    
    \item \textbf{BirdCLEF Benchmarks:} The annual BirdCLEF competitions \cite{kahl2021} provide invaluable datasets for soundscape analysis. However, these datasets exhibit a high degree of domain shift relative to our objectives due to their coverage in species.
\end{itemize}

We compare the list of german birds from the Deutsche Ornithologische Gesellschaft (DOG) against available species within the XC dataset and download bird species audio with appearances in Germany through the XC API. 

\subsubsection{Audio Preprocessing}
\label{sec:audio-preprocessing}
Raw audio recordings are standardized using a multi-stage preprocessing pipeline designed for efficiency and feature consistency. Input recordings in MP3, WAV, or FLAC format are decoded, downmixed to mono, and resampled to a target sampling rate of $f_s = 32{,}000$~Hz. This choice yields a Nyquist frequency of 16~kHz, which is sufficient to capture the high-frequency structure of most passerine vocalizations while reducing computational and storage overhead compared to higher-resolution audio \cite{kahl2021}.

Time-frequency representations are computed as mel spectrograms with 128 mel bands, a 512-point FFT, and a hop length of 512 samples. Spectral magnitudes are converted to a logarithmic decibel scale, clipped to the range $[-80, 0]$~dB, and linearly quantized to 8-bit unsigned integers for compact storage. During training, spectrograms are de-quantized and reconstructed on the original decibel scale.

All spectrograms are temporally standardized to a fixed length of 1000 frames. Shorter sequences are padded using circular (wrap) padding to preserve local temporal continuity, while longer sequences are truncated. Features are then normalized following the PaSST convention using the AudioSet statistics ($\mu = -4.2677393$, $\sigma = 4.5689974$) with an additional scaling factor of two,
\begin{equation}
\hat{x} = \frac{x - \mu}{2\sigma},
\end{equation}
which stabilizes training and aligns the input distribution with the pretrained backbone.

\begin{figure}[t]
    \centering
    \includegraphics[width=\linewidth]{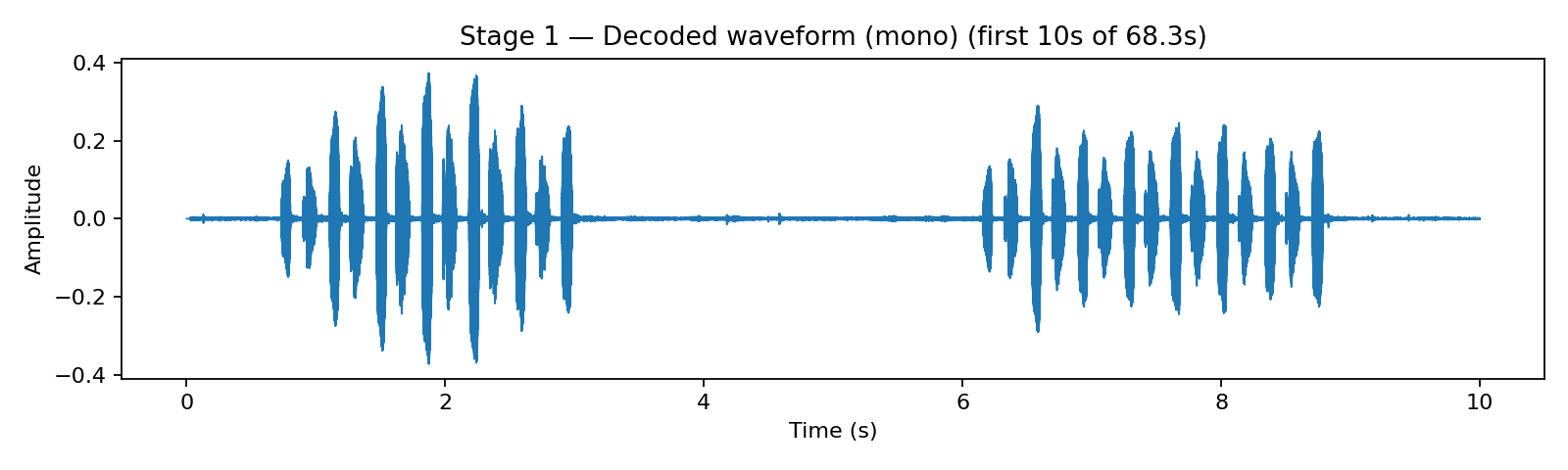}
    \vspace{0.5em}
    \includegraphics[width=\linewidth]{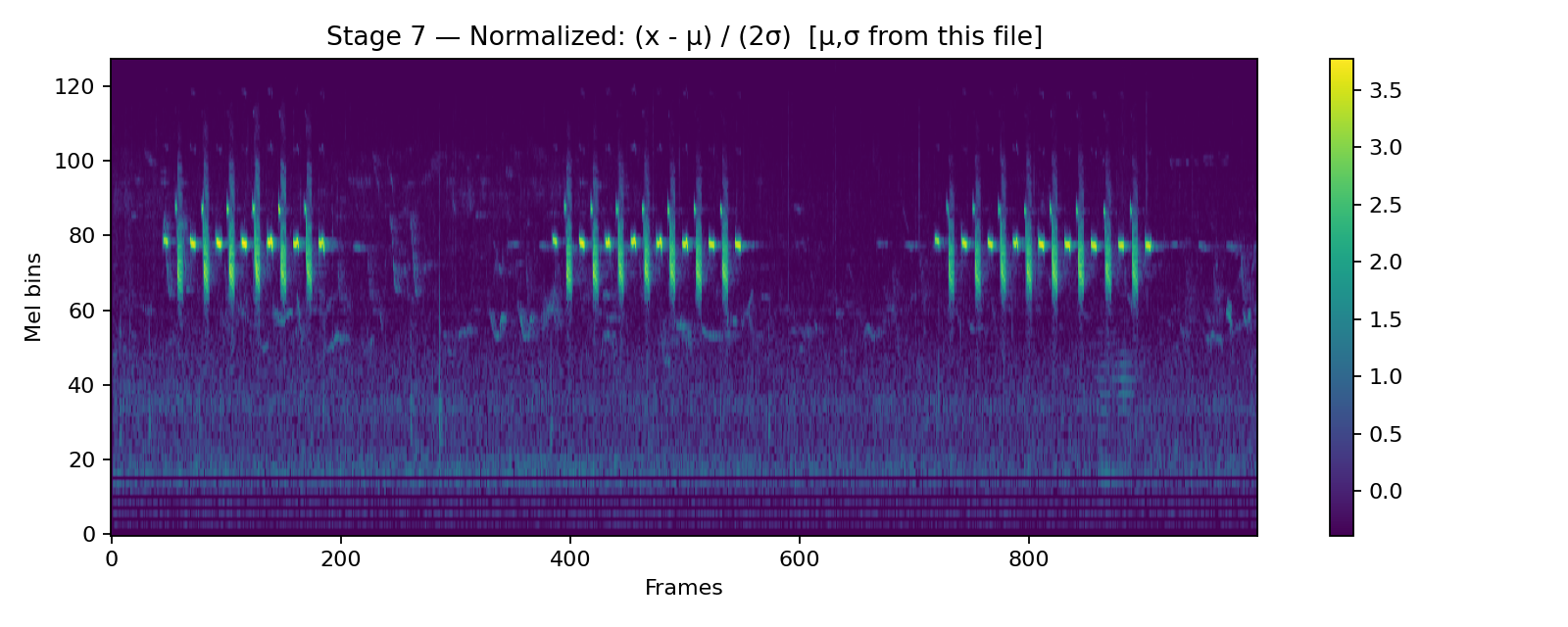}
    \caption{Illustration of the audio preprocessing pipeline.
    \textbf{(top)} Raw mono waveform decoded from the original recording (first 10\,s shown).
    \textbf{(bottom)} Corresponding mel spectrogram after resampling to 32~kHz, log-scaled feature extraction, temporal standardization to 1000 frames, and PaSST-style normalization.
    }
    \label{fig:audio_preprocessing_example}
\end{figure}

To improve generalization, SpecAugment-style data augmentation is applied during training only, consisting of random frequency masking (up to 15 mel bands) and temporal masking (up to 35 frames). Validation samples are processed without augmentation. The PaSST model consumes the resulting single-channel mel spectrograms of shape $1 \times 128 \times 1000$ directly, with its internal audio frontend disabled.

Bioacoustic datasets frequently exhibit a pronounced long-tail distribution, as illustrated in Figure~\ref{fig:class_dist_audio}, where common species dominate recordings while rare species contribute comparatively few samples. To mitigate majority-class bias, we apply stratified oversampling at the dataset level, enforcing a minimum of $500$ samples per species prior to data splitting.

To further address residual imbalance, we employ cost-sensitive learning by computing inverse-frequency class weights $w_c$:
\begin{equation}
    w_c = \frac{N}{C \cdot n_c},
\end{equation}
where $N$ denotes the total number of training samples, $C$ the number of classes, and $n_c$ the sample count for class $c$. The resulting dataset is partitioned into $90\%$ training and $10\%$ validation sets using a stratified split to preserve class distributions across both subsets.

\subsubsection{Bird Activity Detection}
\label{sec:bird-activity-detection}

In continuous monitoring deployments, the vast majority of captured audio contains no avian vocalizations ambient noise, wind, anthropogenic sounds, or silence dominate the recorded signal. Routing every audio segment through a computationally expensive species classifier would incur unnecessary computational overhead on resource-constrained embedded hardware. To address this, we introduce a lightweight binary classification model that serves as a gating mechanism, determining whether a given audio segment contains bird activity before forwarding it to the species classification stage described in Section~\ref{sec:classification-methodologies}.

\paragraph{Training Data}
The training set for the activity detector was derived from two complementary sources. Positive samples (\textit{bird}) were obtained by extracting 3-second segments from Xeno-Canto \cite{vellinga2015} recordings of quality categories A, B, and C - the same corpus used for species classification. Negative samples (\textit{no bird}) were drawn from the ESC-50 dataset \cite{piczak2015esc}, a publicly available collection of 2,000 environmental audio clips spanning 50 categories such as rain, wind, engine noise, footsteps, dog barks, and street traffic. By training against this diverse set of non-avian sounds, the model learns to discriminate bird vocalizations from a representative distribution of real-world background conditions.

\paragraph{Feature Extraction}
Each 3-second audio chunk, sampled at 48\,kHz, is transformed into a log-scaled mel spectrogram using 64 mel-frequency bins, an FFT window size of 2048, and a hop length of 512 samples. The resulting time-frequency representation has shape $64 \times 63 \times 1$ and serves as input to the classification network. Note that the feature extraction parameters of the activity detector differ from those of the species classification pipeline (Section~\ref{sec:audio-preprocessing}), as the binary detection task requires lower spectral resolution and benefits from reduced input dimensionality to minimize inference cost.

\paragraph{Model Architecture}
The bird activity detection model follows a hybrid CNN-GRU architecture designed to jointly capture spectral patterns and temporal dynamics. The convolutional front-end consists of three successive blocks, each comprising a 2D convolutional layer with kernel size $3 \times 3$, batch normalization, ReLU activation, and $2 \times 2$ max pooling. Filter counts increase progressively through the layers (32, 64, 128), enabling the network to learn hierarchical spectro-temporal features--from low-level frequency patterns in early layers to abstract acoustic event representations in deeper layers.

The output of the convolutional stack is reshaped along the temporal axis and passed into two stacked Gated Recurrent Unit (GRU) layers with 64 and 32 hidden units, respectively. The recurrent component models temporal dependencies across the spectrogram's time axis, capturing the characteristic sequential structure of bird calls that distinguishes them from stationary background noise. A dropout rate of $p = 0.3$ is applied after the recurrent layers to mitigate overfitting.

The classification head consists of a single dense neuron with sigmoid activation, producing a probability score $p_{\text{bird}} \in [0, 1]$ representing the likelihood that the audio segment contains bird vocalizations. During inference, a segment is classified as containing bird activity if $p_{\text{bird}} \geq 0.80$.

\paragraph{Deployment}
After training, the model was quantized and converted to TensorFlow Lite format for efficient on-device inference. The resulting model occupies approximately 12\,KB of storage and achieves an inference latency of ${\sim}5$\,ms on the Raspberry~Pi~5, making it suitable for continuous real-time monitoring without significant impact on overall system load. This two-stage design--lightweight activity detection followed by species classification--substantially reduces average computational cost, as the resource-intensive species classifier is invoked only when bird activity is detected. The same gating strategy is employed in the iOS application, where an identical confidence threshold of $p_{\text{bird}} \geq 0.80$ is applied prior to species inference.

\subsubsection{Audio Classification Methodologies}
\label{sec:classification-methodologies}
The classification task is formulated as a multi-class problem covering the $C$ most common bird species in Germany. We evaluate two deep learning approaches previously found within the analysis of related work: Convolutional Neural Networks (CNNs), which serve as the established baseline for bioacoustic classification, and Spectrogram Transformers, representing the current state of the art in audio-based event recognition.

\begin{figure}[ht]
    \centering
    \includegraphics[width=\linewidth]{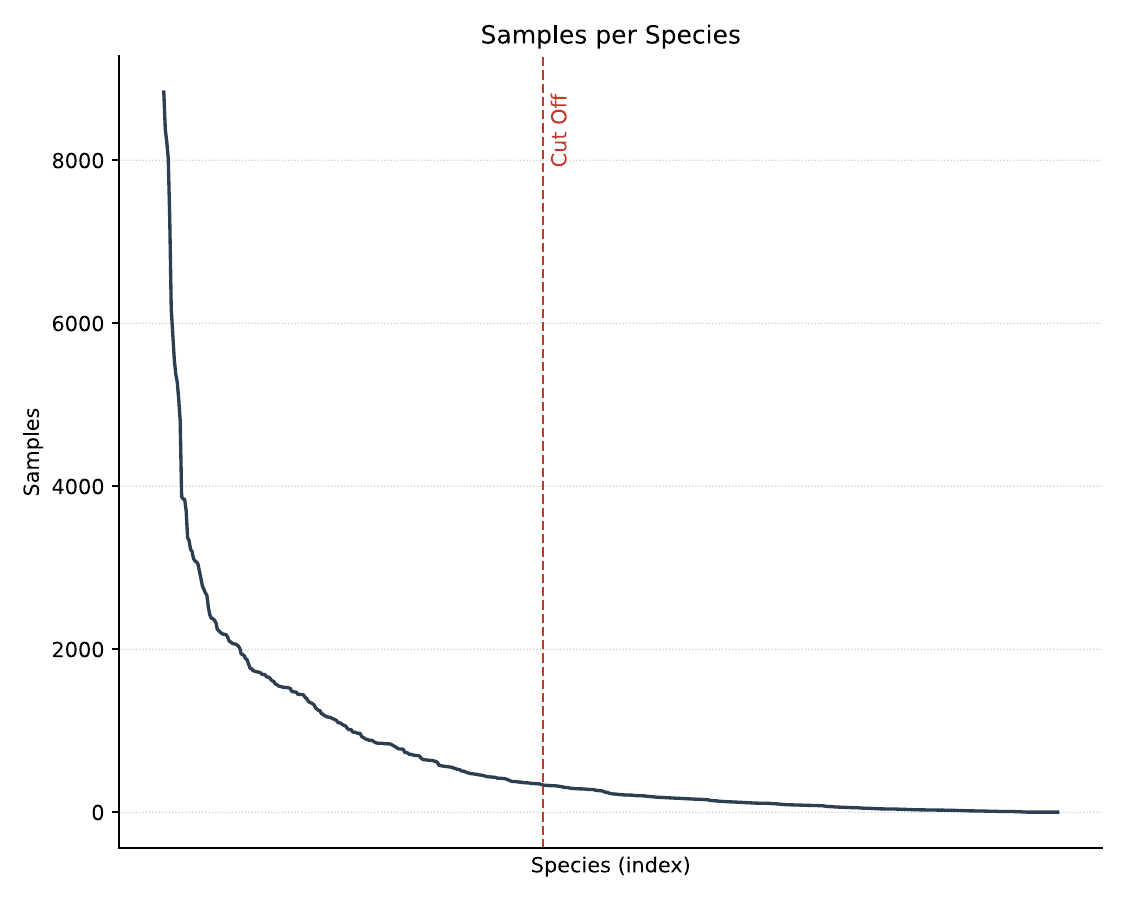}
    \caption{Class distribution of the dataset incl 256 classes cut-off.}
    \label{fig:class_dist_audio}
\end{figure}

To examine the trade-off between the number of species ($C$) and classification accuracy, we conduct an initial ablation study using a MobileNetV3 model. 
Multiple networks are trained on subsets containing the $C$ most frequent bird species from the Xeno-Canto (XC) dataset. The results indicate consistent performance up to $C = 256$ species, with accuracy declining beyond this point. Consequently, we adopt $C = 256$ as the target class count, as it balances species coverage for practical home-use scenarios with stable model performance during training. The cut-off is visualized in the data-overview in figure \ref{fig:class_dist_audio}.

\textbf{MobileNetV3-Small for Edge Optimization}
For scenarios requiring ultra-low latency inference or greatly reduced compute capacity, we employ MobileNetV3-Small \cite{howard2019}.
\begin{itemize}
    \item \textbf{Architecture:} It utilizes inverted residual blocks with Squeeze-and-Excitation (SE) modules. A key distinction is the use of the \textbf{Hardswish} activation function, which approximates the SiLU non-linearity while being computationally cheaper for embedded quantization.
    \item \textbf{Adaptation:} We modified the input stem to accept single-channel spectrograms by replacing the first convolutional layer with $Conv2d(1, 16, 3 \times 3, \text{stride}=2)$. The original classification head is replaced with a regularized block: $Linear_{576 \to 1024} \to Hardswish \to Dropout(0.4) \to Linear_{1024 \to 256}$, where 576 represents the feature dimension after global average pooling over the spatial dimensions of the extracted spectrogram features.
\end{itemize}

\textbf{EfficientNet Family (B0 \& B3)}
We evaluate the EfficientNet family \cite{tan2019} to test the efficacy of \textit{Compound Scaling} in bioacoustics.
\begin{itemize}
    \item \textbf{EfficientNet-B0:} Serving as our primary baseline, B0 balances depth, width, and resolution to minimize FLOPS while maintaining robust feature extraction. We employ label smoothing ($\alpha=0.1$), weight decay ($\lambda=10^{-4}$), and dropout ($p=0.3$) in the classifier to prevent overfitting.
    \item \textbf{EfficientNet-B3:} To investigate if increased model capacity yields better representations of fine-grained bird calls, we scale up to the \textbf{B3} variant. B3 features significantly more layers and wider channels than B0. To mitigate the increased risk of overfitting associated with this larger capacity, we employ a two-stage classifier architecture (1536 $\to$ 1280 $\to$ 640 $\to$ 256) with progressive dropout rates ($p=0.5$ after the first stage, $p=0.3$ after the second stage), combined with label smoothing ($\alpha=0.1$), weight decay ($\lambda=10^{-4}$), and gradient clipping (max norm = 1.0).
\end{itemize}

\textbf{PaSST: Phase-Based Fine-Tuning}

We initialize PaSST with weights pre-trained on AudioSet and apply a two-stage fine-tuning protocol optimized for avian vocalizations:

\begin{itemize}
    \item \textbf{Phase 1 (Linear Probing):} The Transformer backbone is frozen for 20 epochs. Only the classification head (768 $\to$ 1024 $\to$ GELU $\to$ Dropout $\to$ $n_\text{classes}$) is trained with learning rate $\text{LR} = 1 \times 10^{-4}$.
    
    \item \textbf{Phase 2 (Full Fine-Tuning):} The entire backbone is unfrozen for 30 epochs with reduced learning rate $\text{LR} = 1 \times 10^{-5}$. We employ Mixup augmentation ($\alpha=0.4$), weighted Cross-Entropy loss with label smoothing ($\epsilon=0.1$), AdamW optimization ($\lambda_{\text{wd}}=10^{-4}$), OneCycleLR scheduling, gradient accumulation (16 steps), and gradient clipping (max norm = 1.0). Checkpoints are saved based on best validation top-1 accuracy.
\end{itemize}

\subsubsection{Comparative Classification Benchmark}

\begin{table}[ht]
\centering
\caption{Model Performance Summary (Best Validation Accuracy)}
\label{tab:model_comparison}
\begin{tabular}{lccc}
\hline
\textbf{Model} & \textbf{Top-1 Accuracy (\%)} & \textbf{Top-5 Accuracy (\%)} \\ \hline
PaSST                            & 94.39                        & 97.60                        \\
EfficientNetB3                   & 92.93                        & 97.37                        \\
EfficientNetB0 & 91.69                        & 97.31                        \\
MobileNetv3                     & 85.62                        & 94.75                        \\ \hline
\end{tabular}
\end{table}

\begin{figure}[ht]
    \centering
    \includegraphics[width=\linewidth]{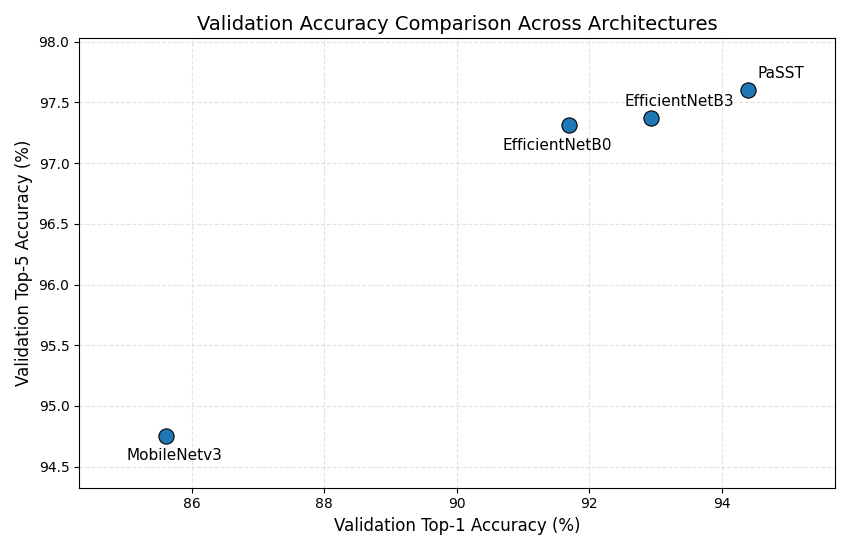}
    \caption{Accuracy comparison for audio classification architectures on 256 bird species.}
    \label{fig:modelbenchmark}
\end{figure}

We evaluate four architectures on the bird species classification task as shown in table \ref{tab:model_comparison}. \textbf{PaSST} achieves the highest top-1 validation accuracy at 94.39\%, outperforming EfficientNetB3 (92.93\%), EfficientNetB0 (91.69\%), and MobileNetv3 (85.62\%) by 1.46, 2.70, and 8.77 percentage points respectively. Top-5 accuracy is tighter across architectures (94.75\%-97.60\%), indicating all models reliably rank the correct species within top-5 predictions (Figure \ref{fig:modelbenchmark}).

Transformer-based self-attention mechanisms prove superior for capturing long-range temporal dependencies in avian vocalizations compared to convolutional inductive biases. Notably, EfficientNet variants converge faster (epoch 11) than PaSST (epoch 27), suggesting that CNNs align more naturally with spectral features, while Transformers require extended fine-tuning to fully adapt pre-trained AudioSet representations. EfficientNetB3 demonstrates the efficacy of compound scaling over B0 (1.24\% improvement), yet the performance gap from PaSST highlights that architectural choice outweighs model capacity for this domain. PaSST is recommended for deployment where computational budget permits, while EfficientNetB0 offers a practical baseline (91.69\%) for resource-constrained edge devices.

\subsection{Visual bird classification}
Visual bird species recognition is addressed using two alternative modeling strategies introduced earlier in this work. As images may contain multiple birds per scene, the task is formulated beyond standard single-label image classification. We evaluate (i) a multiclass object detection approach that jointly predicts bird locations and species labels, and (ii) a two-stage pipeline in which bird instances are first localized and subsequently classified using a dedicated fine-grained image classifier. The following sections describe the data preparation and classification methodologies for both approaches.

\subsubsection{Data Selection and Preprocessing}
The first step of this work was to identify an existing image dataset suitable for fine-grained bird species classification.
The primary requirement was that the dataset must contain all bird species that may occur in Germany, including migratory and rare species. Therefore, this project requires species-level discrimination across several hundred taxa \cite{gedeon2014}. Additionally, species names had to be taxonomically consistent with the Xeno-Canto database, which is used as the audio data source in the multimodal classification pipeline. Xeno-Canto follows a strict ornithological taxonomy based on current scientific consensus, making taxonomic alignment a critical constraint \cite{vellinga2015}.

As the visual classification stage was implemented after the audio-based species filtering, the number of required visual classes was reduced to 256.
An initial taxonomic mapping between Xeno-Canto species names and iNaturalist taxa was performed.

Out of all target species, three could not be found in iNaturalist.
\textit{Acanthis cabaret} is treated as a subspecies (\textit{Acanthis flammea cabaret}) rather than a distinct species in the iNaturalist taxonomy and is therefore not listed separately.
\textit{Anser cygnoid f. domestica} and \textit{Columba livia f. domestica} represent domesticated forms.
iNaturalist focuses on wild organisms and explicitly excludes domesticated or cultivated forms from most taxonomic datasets; consequently, these taxa are outside the scope of iNaturalist and were excluded from further analysis.

Several publicly available bird image datasets were evaluated, including CUB-200-2011 \cite{wah2011cub} and other regional datasets.
While these datasets are widely used in computer vision research, they do not fulfill the requirement of containing primarily European bird species.
To the best of our knowledge, no publicly available dataset provides comprehensive image coverage for all German bird species at the species level.

Given these limitations, the iNaturalist platform was selected as the source for image data.
iNaturalist is a large-scale citizen science project that provides millions of georeferenced biodiversity observations, including photographs annotated with taxonomic information and community-based verification \cite{vanhorn2018inat,inaturalist2024}.
Primarily, images that were verified as \textit{research grade} were selected.
For some species, the number of available research-grade images was limited; therefore, in a second approach, all available observations were considered.

The public iNaturalist API was used to collect the data, enabling reproducible data acquisition.
All images were filtered to include Creative Commons–licensed content (CC0, CC-BY, CC-BY-SA). A small subset (approximately 2\%) of the downloaded images did not provide explicit license information at the time of access and were therefore marked as \textit{unknown}. These cases were retained for analysis and explicitly documented. Due to licensing constraints, the image data used in this study cannot be redistributed directly. For reproducibility, the metadata files and the corresponding download scripts are provided within the accompanying GitHub repository.

The data preprocessing has to be considered individually for each approach. First, the preprocessing steps for the multiclass object detection approach are described, followed by those for the two-stage pipeline.

\paragraph{Multiclass Object Detection}  
For the multiclass object detection approach, a weakly supervised annotation strategy was employed to generate bounding box annotations from image-level species labels.
The original dataset consisted of species-wise organized images obtained from iNaturalist, where each image is associated with a single bird species label but lacks explicit spatial annotations.

To derive bounding boxes, a pretrained YOLOv11 object detector \cite{ultralytics_yolov11}, building on the YOLO family of detectors \cite{redmon2016yolo}, was used. 
For each image, the detector was applied with a low confidence threshold to maximize recall.
Only detections corresponding to the COCO bird class were considered, and in cases where multiple detections were present, the bounding box with the highest confidence score was retained.
This top-1 selection reduces annotation noise in images containing multiple birds or background clutter.

The selected bounding box was subsequently assigned the image-level species label, yielding a multiclass detection annotation where each bounding box corresponds to a specific bird species.
Bounding boxes were slightly expanded by a fixed padding factor to ensure full object coverage and to mitigate tight cropping effects.
Very small detections relative to the image area were discarded to avoid spurious annotations.

Images for which no bird could be detected were excluded from the detection dataset.
A limited number of such failure cases were retained for qualitative inspection to assess detector behavior.
The resulting annotations were stored in YOLO format using normalized center coordinates and width–height representations.

The final detection dataset was split into training, validation, and test subsets.
While the original data provided only training and validation splits, an additional test set was created by stratified sampling from the training data to preserve class balance.
This procedure ensures a fair evaluation while avoiding information leakage across splits.

It should be noted that this weakly supervised labeling process inherently introduces a degree of noise, particularly in images containing multiple individuals or visually ambiguous species.
However, similar strategies have been shown to be effective for large-scale ecological vision tasks where manual annotation is infeasible \cite{vanhorn2018inat, norouzzadeh2018automatically}.

\paragraph{Two-Stage Detection–Classification Pipeline}  
In the two-stage detection–classification pipeline, object localization and species recognition are explicitly decoupled.
A pretrained object detector is first used to localize birds within the full-resolution images. The resulting object-centric image crops are then used as inputs for a dedicated fine-grained species classifier.

To generate training data for the classification stage, bounding-box–based pseudo-crops were created using a pretrained YOLO detector \cite{ultralytics_yolov11}. For each image, the detector was applied in inference mode and restricted to the generic COCO class bird  (class index 14). To reduce ambiguity in scenes containing multiple individuals, only the most confident bird detection per image was retained. Detections with a confidence score below 0.1 or bounding boxes covering less than 0.5\% of the original image area were discarded. The non-maximum suppression IoU threshold was set to 0.5. 

Applying this filtering reduced the dataset from 114,785 original images to 97,325 images. This reduction reflects the characteristics of iNaturalist imagery, which frequently depicts realistic scenes containing complex backgrounds, partial views of birds, or non-target content. Excluding images without a confidently detected bird was therefore necessary to construct a reliable dataset for fine-grained bird species classification.

\begin{figure}[ht]
    \centering
    \includegraphics[width=\linewidth]{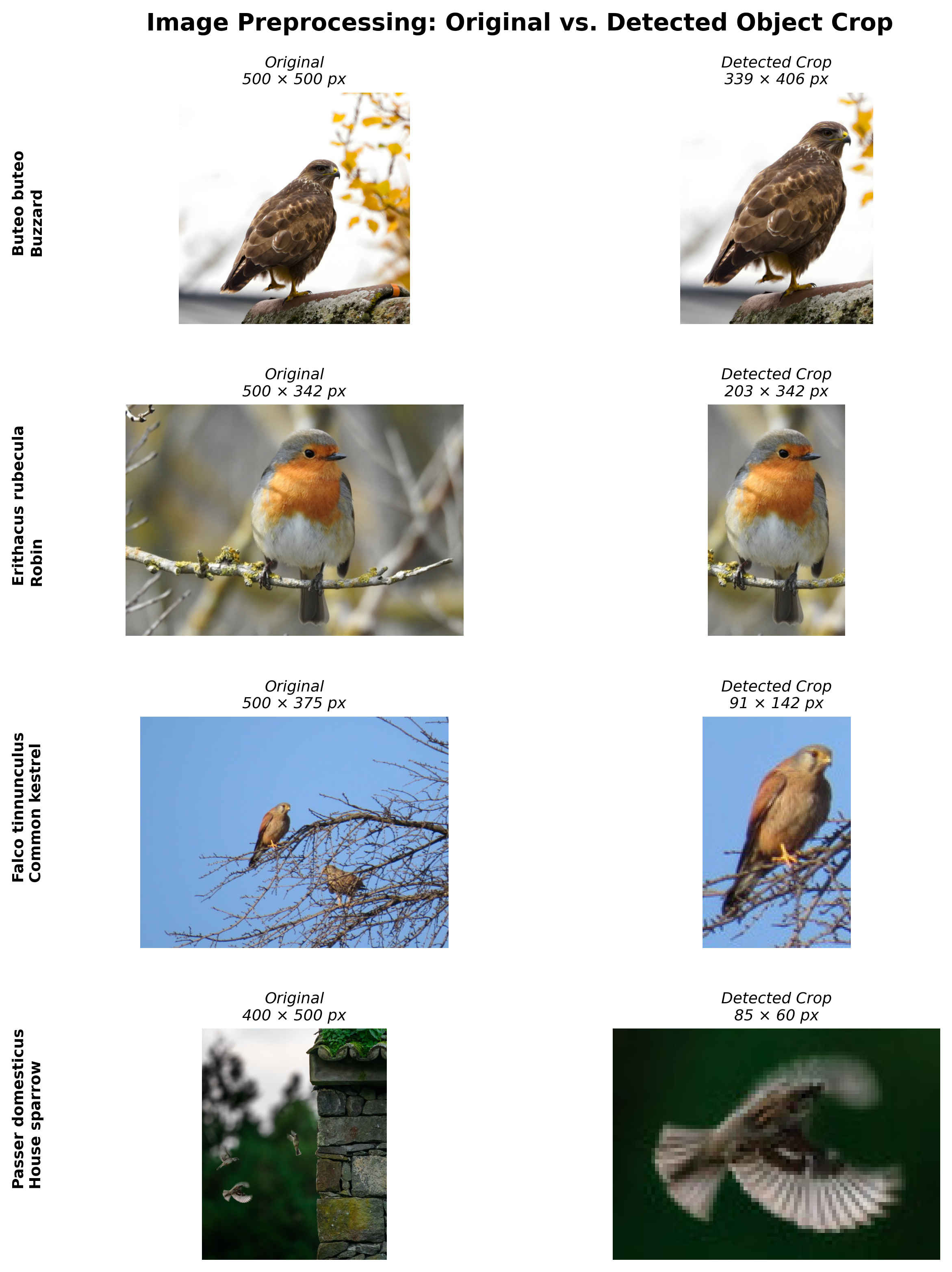}
    \caption{Illustration of the preprocessing pipeline using YOLOv11-based bird detection followed by image cropping. Four example images sourced from iNaturalist \cite{inaturalist2024} are shown, demonstrating varying original image sizes and resulting crops:\\
\textit{Buteo buteo} (Observation ID: 325504686, Photo ID: 590714713, CC-BY),\\
\textit{Erithacus rubecula} (Observation ID: 331794775, Photo ID: 601986829, CC-BY),\\
\textit{Falco tinnunculus} (Observation ID: 322740349, Photo ID: 583899233, CC0),\\
\textit{Passer domesticus} (Observation ID: 333737125, Photo ID: 605910979, CC0).\\
Bounding boxes and cropped images were generated by the authors.}
    \label{fig:preprocessing_visualization_1}
\end{figure}

The selected bounding boxes were expanded by a fixed padding factor of 15\% to preserve contextual visual cues such as wings, tail feathers, and surrounding habitat. Each cropped image was optionally resized to a maximum side length while preserving the original aspect ratio. Consequently, the generated crops exhibit variable spatial resolutions, with minimum side lengths ranging from 29 to approximately 500 pixels. Figure~\ref{fig:preprocessing_visualization_1} illustrates this preprocessing pipeline and highlights the necessity of bird detection, as target objects may appear at arbitrary positions within the original images.

Importantly, a fixed input resolution was not enforced during crop generation, but instead during data loading and augmentation. All images were transformed to a standardized resolution of 224×224 pixels within the training and validation pipelines. During training, a RandomResizedCrop(224) operation was applied, sampling between 40\% and 100\% of the available crop area and upsampling smaller images as necessary. For validation, images were resized to 255 pixels on the shorter side, followed by a center crop of 224×224 pixels. This guarantees that all inputs fed into the classification models have identical spatial dimensions, independent of the original crop size.

Although this procedure involves upsampling for a substantial fraction of the crops, modern convolutional neural networks pretrained on large-scale natural image datasets such as ImageNet are known to learn a considerable degree of scale invariance. Prior work has shown that such pretrained CNNs retain robust semantic representations across variations in object scale and effective image resolution, which supports their use in transfer learning scenarios with heterogeneous input resolutions \cite{graziani2021scale}. In addition, extensive data augmentation, including RandAugment, color jittering, random erasing, and CutMix-further mitigates potential artifacts introduced by interpolation. The resulting training distribution comprises a mixture of high-resolution and low-resolution bird-centric crops.

All preprocessing steps were performed in a weakly supervised manner, as no manually annotated bounding boxes were available. Consequently, the resulting crops may contain a degree of label noise, particularly in images with multiple birds or partial occlusions. However, similar pseudo-labeling strategies have been shown to be effective for large-scale ecological datasets where exhaustive manual annotation is infeasible \cite{vanhorn2018inat, norouzzadeh2018automatically}.

\paragraph{Dataset distribution and constraints for the two-staged pipeline}

After cropping, the dataset was reorganized into a clean, stratified train/validation/test split. A species-wise 60/20/20 partitioning was applied to ensure that each species was represented across all evaluation subsets whenever sufficient data were available. This restructuring enables a fair and reproducible evaluation of the classification models on a held-out test set while preserving the inherent class distribution of the data.

\begin{figure}[ht]
    \centering
    \includegraphics[width=\linewidth]{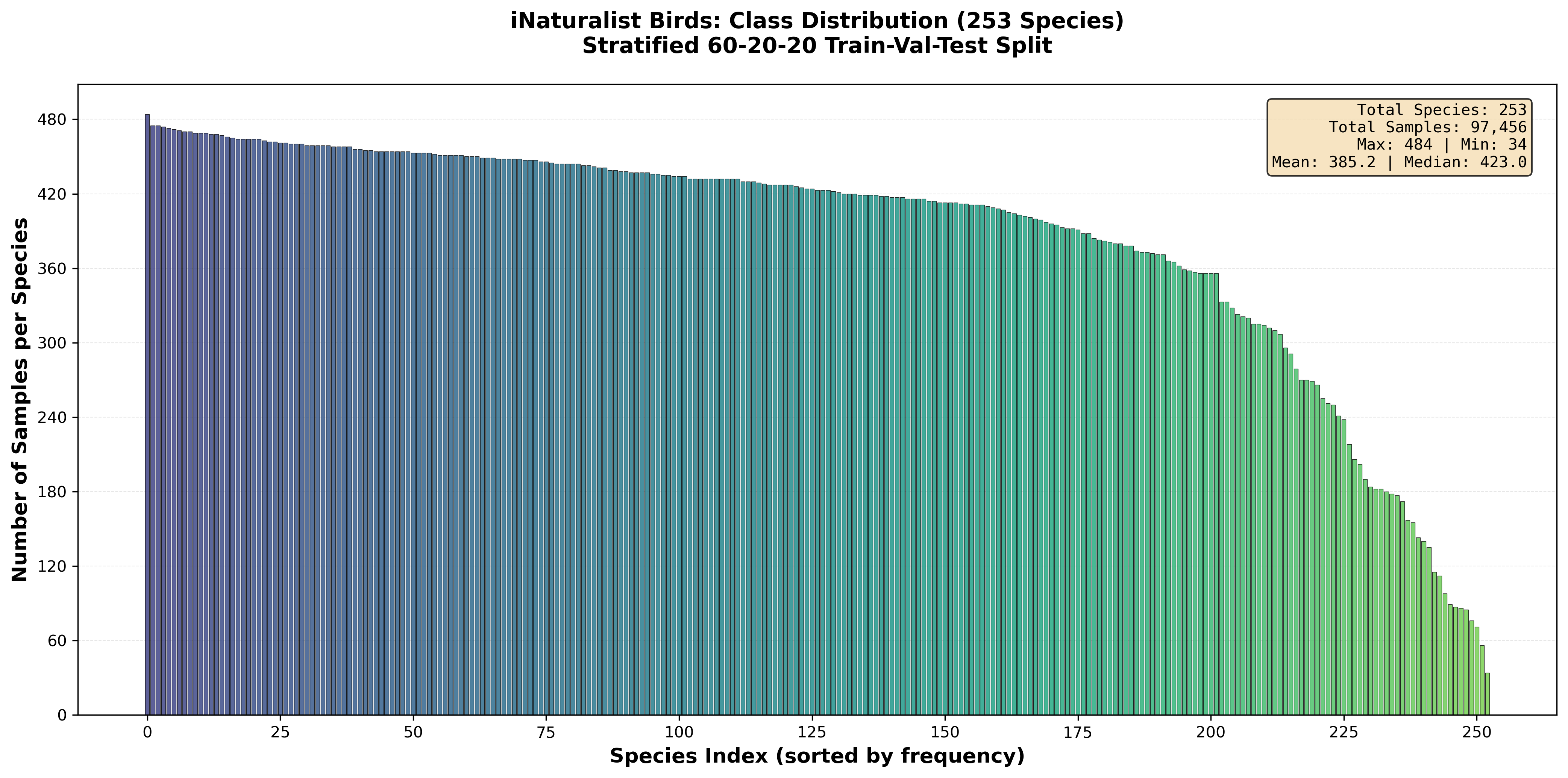}
    \caption{Class distribution of the dataset after object detection and crop generation for the two-stage pipeline.}
    \label{fig:class_distribution_2stage}
\end{figure}

Figure~\ref{fig:class_distribution_2stage} illustrates the resulting number of images per species after object detection and crop generation. As described above, for species with fewer than 500 available research-grade images, all usable samples were retained to maximize data coverage. Consequently, the resulting dataset exhibits a pronounced class imbalance, reflecting the natural observation frequency and reporting bias present in large-scale citizen science datasets such as iNaturalist.

No additional balancing strategies were applied in this study. While such methods may further improve performance for underrepresented species, addressing dataset imbalance was considered beyond the scope of the present work.

\subsection{Classification Pipelines}

This section describes the two visual classification pipelines evaluated in this work: a single-stage multiclass object detection model and a two-stage detection-classification pipeline.

\subsubsection{Multiclass Object Detection Pipeline}
The multiclass object detection pipeline is implemented using the Ultralytics YOLO framework. The detector was initialized from a pretrained YOLO26 model (configuration: \texttt{yolo26s.pt}), whose weights were obtained from large-scale pretraining on the COCO dataset \cite{sapkota2026yolo26keyarchitecturalenhancements}. The model was subsequently fine-tuned on the customized iNaturalist bird dataset introduced in the preceding sections, using a dataset specification provided via a YOLO-compatible YAML file.
Given the weakly supervised nature of the annotation process and the hierarchical folder layout, the training setup included a lightweight dataset integrity verification prior to optimization. In particular, label files were checked recursively to account for class-wise subdirectories, and a small sample of annotations was parsed to verify the expected YOLO label format (class index and normalized bounding box parameters). This sanity check aimed to identify empty label files and malformed entries early, thereby reducing the risk of silent training failures and improving the reliability of the subsequent evaluation.
Training was performed with the Ultralytics high-level API (\texttt{model.train}) using a fixed input resolution of $640 \times 640$ pixels and a batch size of 32. The compute device was selected automatically, prioritizing CUDA acceleration when available, followed by MPS or CPU execution. Data loading employed multiple workers to mitigate input bottlenecks. A fixed random seed was applied to stabilize the experimental setup and support reproducibility at the pipeline level.
Optimization was configured via Ultralytics’ optimizer selection (\texttt{optimizer="auto"}) that uses the with YOLO26 introduced MuSGD optimizer \cite{sapkota2026yolo26keyarchitecturalenhancements}. The initial learning rate was set to $2\times10^{-3}$ with a final learning rate factor of 0.01, momentum of 0.937, and weight decay of $10^{-3}$. A warmup phase of two epochs was used (including warmup momentum and bias learning rate adjustments), and cosine learning-rate scheduling was enabled to promote stable convergence over long training horizons. Training was run for up to 150 epochs with early stopping controlled by a patience of 30 epochs. For efficiency, mixed-precision training (AMP) was enabled and samples were cached in RAM.
Generalization was supported through controlled data augmentation tailored to fine-grained bird detection. Mosaic augmentation was applied during early training (mosaic probability 0.2) and disabled after 10 epochs to stabilize later-stage optimization \cite{bochkovskiy2020yolov4optimalspeedaccuracy}. Additional augmentations included modest translation and scaling, horizontal flipping, and HSV-based color jitter, while more aggressive geometric distortions (e.g., rotation, shear, perspective) were disabled. Loss contributions were explicitly balanced using increased box regression weight, moderate classification and distribution focal loss weights, and mild label smoothing.
Each training run produced standard Ultralytics artifacts, including periodic checkpoints and final weights, learning curves, and structured logs which are visualized in figure 6. After training, the best-performing checkpoint was evaluated on the validation split with a low confidence threshold (0.001) and an IoU threshold of 0.7, using a capped number of detections per image. Performance was reported via standard detection metrics (precision, recall, F1, mAP@0.5, mAP@0.75, and mAP@0.5:0.95), with optional per-class analyses. In addition, an evaluation on the training split was performed as an overfitting diagnostic, and metric summaries were exported in machine-readable form to facilitate downstream analysis and reporting.
\begin{figure}[ht]
    \centering
    \includegraphics[width=\linewidth]{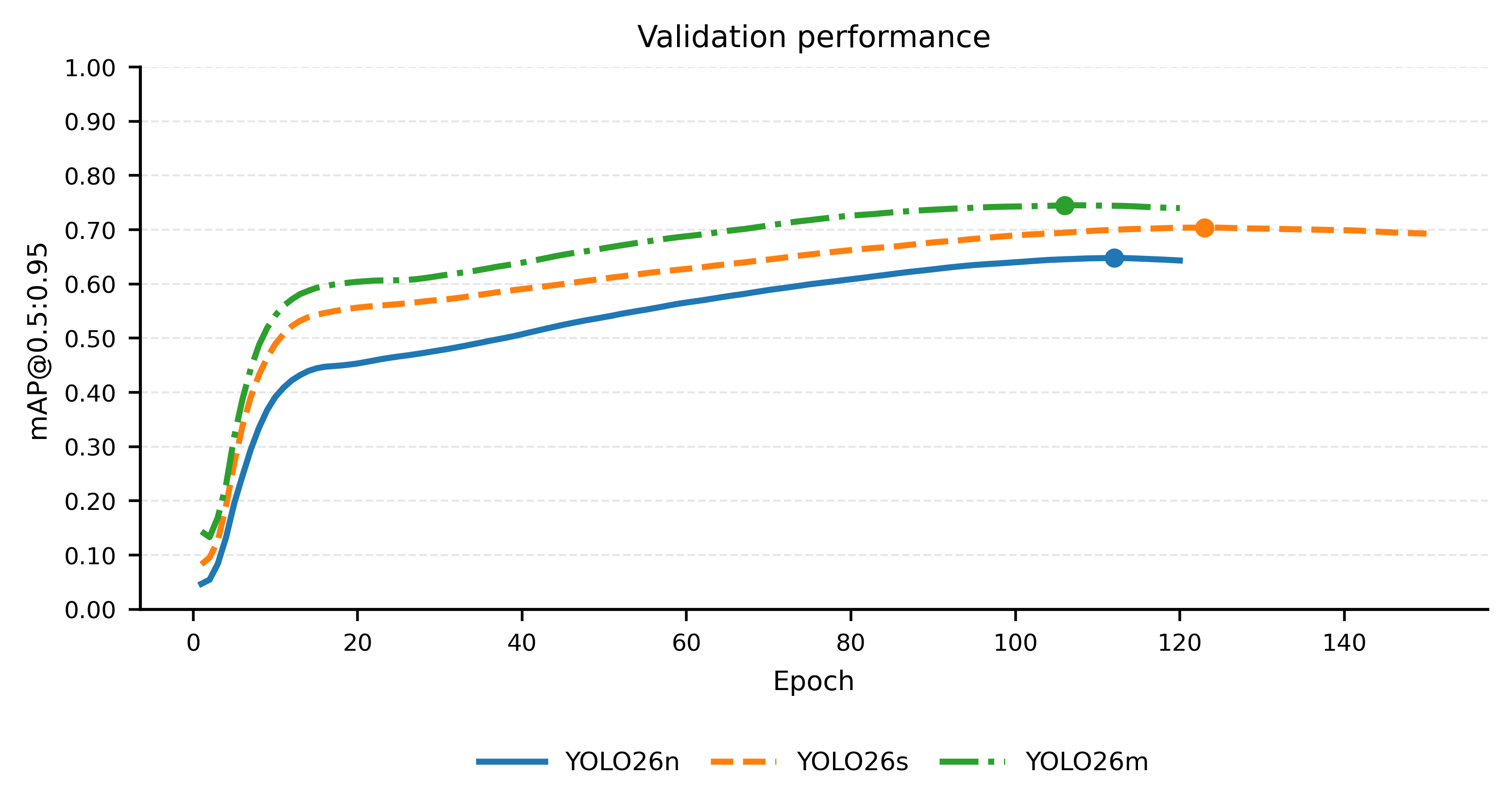}
    \caption{YOLO26 validation results during training compared for YOLO26n, YOLO26s and YOLO26m}
    \label{fig:validation_visualization}
\end{figure}

\textbf{Training results Visual Species Classification}

The evaluation results for the various model configurations on the held-out test split (n=19,593) are summarized in Table \ref{tab:visual_results_audio}. A consistent positive correlation between model scale and detection accuracy is evident across all reported metrics. The baseline YOLO26n variant achieves an mAP@0.5 of 0.7423, while scaling to the small (YOLO26s) configuration yields an improvement of approximately 4.8 percentage points, reaching 0.7908. The peak performance is observed with the YOLO26m model, which attains an mAP@0.5 of 0.8224 and a rigorous mAP@0.5:0.95 of 0.7442.

These results suggest that the increased parameter capacity of the medium variant is effectively utilized to differentiate the fine-grained morphological features of the various bird species. Notably, these performance levels were achieved despite the challenges inherent in the weakly supervised labeling process. This robustness is attributed to the specialized training pipeline, which promoted stable convergence through cosine learning-rate scheduling and the strategic deactivation of mosaic augmentation during the final training phases (close-mosaic). The high mAP@0.5:0.95 values further indicate that the models not only correctly classify the species but also maintain high localization precision.

\begin{table}[t]
\centering
\caption{Test set performance of visual species classification models. All metrics are reported on the held-out test split (19,593 samples).}
\label{tab:visual_results_audio}
\begin{tabular}{lcccccc}
\hline
\textbf{Model} & \textbf{mAP@0.5} & \textbf{mAP@0.5:0.95}\\
\hline
YOLO26n & 0.7423 & 0.6517 \\
YOLO26s   & 0.7908 & 0.7051 \\
YOLO26m   & \textbf{0.8224} & \textbf{0.7442}\\
\hline
\end{tabular}
\end{table}

\subsubsection{Two-Stage Detection--Classification Pipeline}

The detection stage of the proposed two-stage pipeline has been described in detail as part of the dataset preprocessing. In the following, we focus exclusively on the image-based species classification stage.

The resulting object-centric crops exhibit substantial variability in spatial resolution. Instead of enforcing a minimum crop size during preprocessing, input size consistency was ensured during training via the transformation pipeline. All samples were dynamically resized and cropped to a fixed resolution of $224 \times 224$ pixels using split-specific transformations for training and validation. This approach allows the model to accommodate heterogeneous input resolutions while remaining fully compatible with ImageNet-pretrained backbones.

For visual species classification, we report Top-1 Accuracy as the primary performance metric. To account for class imbalance inherent in large-scale biodiversity datasets, we additionally report Balanced Accuracy, defined as the average recall across all classes.

Furthermore, we report class-mean Average Precision (cmAP), computed by averaging the per-class average precision independently for each species. This metric captures ranking quality and confidence calibration, complementing accuracy-based measures.

\subsubsection{MobileNetV3-Small (Edge Optimization)}

For resource-constrained visual inference scenarios, we employ \textbf{MobileNetV3-Small}, consistent with the audio classification pipeline. The architecture remains unchanged, leveraging inverted residual blocks with integrated Squeeze-and-Excitation modules for efficient channel-wise feature recalibration, optimized for mobile and embedded hardware \cite{howard2019}.

The network input stem was adapted to accept three-channel RGB images, and the final classification head was replaced to match the number of target bird species.

\subsubsection{EfficientNet Family (B0 \& B1)}

To evaluate the impact of increased representational capacity in visual fine-grained classification, we employ two variants of the EfficientNet family \cite{tan2019}. EfficientNet-B0 serves as the primary baseline, providing a balanced trade-off between network depth, width, and input resolution with minimal computational overhead.

Additionally, EfficientNet-B1 is evaluated to assess whether moderate compound scaling improves discrimination between visually similar bird species. Compared to B0, the B1 variant jointly increases network depth, channel width, and input resolution while maintaining a manageable computational footprint.

\subsubsection{Training results and model comparison}

To ensure fair model comparison and stable convergence, all classification models were trained under a unified training protocol. A stratified 60/20/20 train/validation/test split was used consistently across all experiments. Training was conducted for a fixed number of epochs, with early stopping applied based on validation performance to mitigate overfitting.

Rather than performing an exhaustive hyperparameter optimization, a limited set of targeted configurations was evaluated. These configurations were informed by prior experiments with MobileNet architectures and refined empirically to assess model sensitivity to key training parameters. The explored factors included learning rate, batch size, weight decay, label smoothing, and the strength of data augmentation techniques such as RandAugment, CutMix, and Random Erasing.

For MobileNetV3-Small, this exploratory process revealed that smaller batch sizes combined with moderately increased learning rates and stronger data augmentation consistently yielded superior validation performance. EfficientNet-B0 benefited from slightly more conservative learning rates and higher batch size, while EfficientNet-B1 achieved its best performance under a training configuration comparable to that of MobileNetV3-Small.

\begin{figure*}[t]
    \centering
    \includegraphics[width=0.9\textwidth]{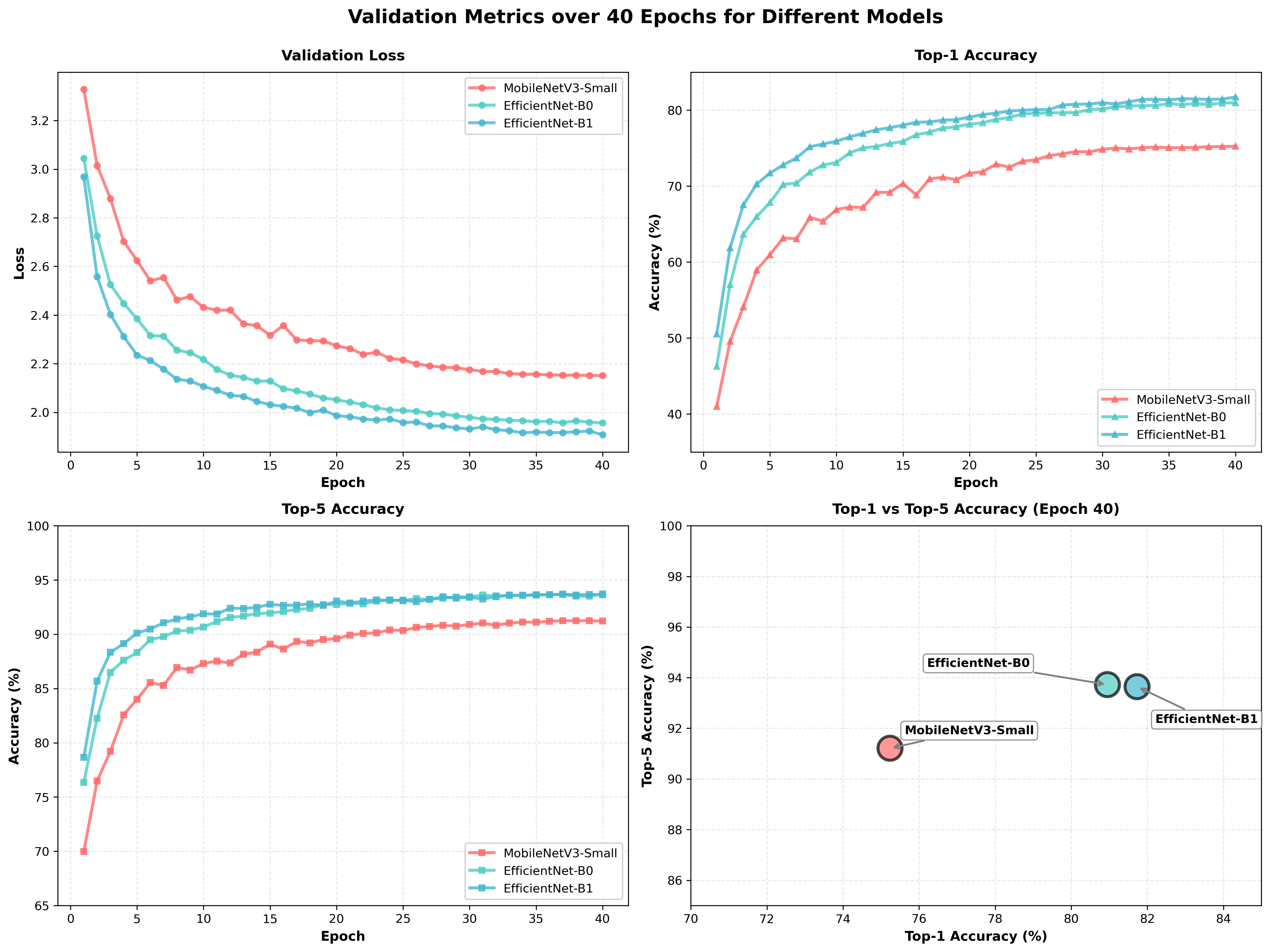}
    \caption{Validation loss, Top-1 accuracy, and Top-5 accuracy during training for all three models.}
    \label{fig:training_curves}
\end{figure*}

All experiments were tracked using MLflow, enabling systematic logging of configurations, metrics, and model checkpoints. For each architecture, the best-performing configuration was selected based on validation top-1 accuracy and subsequently evaluated on the held-out test set. In figure \ref{fig:training_curves}, the validation results during training are compared, showing the same trend observed for audio classification: EfficientNet models achieve better performance than MobileNetV3-Small.

All reported quantitative evaluation results are obtained on the held-out test set and correspond to the best-performing configuration for each architecture, selected based on validation top-1 accuracy.

Table~\ref{tab:visual_results} summarizes the performance of MobileNetV3-Small, EfficientNet-B0, and EfficientNet-B1 across a range of complementary evaluation metrics. These metrics were chosen to capture both overall classification accuracy and robustness in the presence of class imbalance.

\begin{table}[t]
\centering
\caption{Test set performance of visual species classification models. All metrics are reported on the held-out test split (19,593 samples).}
\label{tab:visual_results}
\begin{tabular}{lcccccc}
\hline
\textbf{Model} & \textbf{Top-1 Acc.} & \textbf{Balanced Acc.} &  \textbf{cmAP} \\
\hline
MobileNetV3-Small & 0.7607 & 0.7452 &  0.7188 \\
EfficientNet-B0   & 0.8155 & 0.8007 &  \textbf{0.7637} \\
EfficientNet-B1   & \textbf{0.8206} & \textbf{0.8069}  & 0.7450 \\
\hline
\end{tabular}
\end{table}

EfficientNet-B1 achieves the highest top-1 accuracy of 82.06\%, outperforming both EfficientNet-B0 and MobileNetV3-Small. This improvement indicates that moderate compound scaling of network depth and width yields tangible benefits for fine-grained bird species classification, even when input resolution is kept constant at $224 \times 224$ pixels. 

EfficientNet-B0 demonstrates competitive performance, achieving a slightly lower top-1 accuracy of 81.55\%, while attaining the highest cmAP score among the evaluated models. This indicates strong ranking performance and well-calibrated confidence estimates, despite marginally lower absolute classification accuracy.

MobileNetV3-Small, while substantially more lightweight, achieves a top-1 accuracy of 76.07\%. These results confirm MobileNetV3-Small as a viable option for resource-constrained deployment scenarios where inference efficiency is prioritized over peak accuracy.

\subsection{User hardware deployment}

We explore multiple deployment strategies for home-based and do-it-yourself (DIY) applications of the previously introduced models. These strategies include implementation on continuously operating embedded platforms, such as Raspberry Pi–class single-board computers, as well as integration into a native iOS smartphone application.

\subsubsection{iOS App Development}

The trained PyTorch models were converted to Apple's native Core ML format for deployment. While a direct integration of ONNX models is possible, Core ML was selected to leverage the device's dedicated hardware acceleration (Apple Neural Engine and GPU). This native implementation reduces inference latency and minimizes battery consumption, which is critical for the continuous processing required in the audio module.

Replicating the Python-based feature extraction on iOS presented a challenge, specifically regarding the generation of mel-spectrograms. To align the environments, the exact mel filter banks were extracted from the Python pipeline and implemented in Swift. However, an identical reproduction of the spectrograms is not technically feasible due to differences in the underlying FFT implementations-Python relies on NumPy/MKL, whereas iOS uses the \texttt{Accelerate} framework. These variations in floating-point precision and FFT algorithms lead to minor discrepancies in the input tensors. Consequently, there is a risk that the model performance on the device deviates slightly from the validation results obtained in Python.

\paragraph{Application Workflow}
The user interface provides two input modalities: audio (``Listen'') and image (``Watch'').

\begin{itemize}
    \item \textbf{Image Classification:} Images are captured via the device camera or selected from the photo library and classified by the vision model. Predictions are displayed only when the confidence exceeds $p > 0.2$.
    \item \textbf{Audio Classification:} Audio is recorded continuously and processed in 5\,s segments using a two-stage pipeline. First, an intent model detects bird vocalizations ($p > 0.8$). If detected, a species classifier is applied, with results shown only for probabilities $p > 0.15$.
\end{itemize}

Successful detections are stored locally in a ``Discoveries'' menu. This interface allows users to review captured images and replay audio recordings. Each entry is linked to metadata, including ornithological information and the geolocation of the recording.

\subsubsection{Web Application}
\label{sec:hardware-prototype}

To enable real-time and continous monitoring of classification results, the system provides a lightweight, browser-based dashboard hosted directly on the user hardware such as the Raspberry~Pi or Rubik~Pi. The dashboard follows a client-server architecture in which a minimal HTTP backend serves both a static single-page interface and a RESTful API for asynchronous data access.

Detection events produced by the audio and vision pipelines are aggregated in a shared, thread-safe buffer and exposed through periodic API queries. The frontend polls the backend at fixed intervals to retrieve updated detection records and system statistics, enabling near–real-time visualization while remaining decoupled from the inference processes. Each detection record contains the predicted species, confidence score, timestamp, and input modality.

The dashboard presents system status information and a chronological detection log, allowing users to distinguish between audio- and image-based classifications and to monitor activity over time. Hosting the interface locally on the embedded device enables access from any browser within the home network without requiring command-line interaction or specialized software.

Overall, this design separates inference from visualization with minimal system overhead, supporting accessible, do-it-yourself biodiversity monitoring on resource-constrained edge devices.

In addition to the frame-based visual pipeline, an experimental live video streaming mode was implemented using \texttt{GStreamer}. This mode performs continuous object detection using the YOLO model directly on the video stream, enabling real-time visual feedback. While functionally viable, this approach is severely constrained by the limited computational resources of the Raspberry~Pi and Rubik~Pi, resulting in low frame rates and increased latency.

More generally, although all models employed in this work are capable of running entirely on Raspberry~Pi–class hardware, sustained real-time performance remains challenging, particularly for continuous visual inference. These limitations highlight the trade-off between model complexity and responsiveness on resource-constrained edge devices and motivate the use of low-duty-cycle or event-driven inference strategies.


\section{Conclusion and Future Work}

This work presented \emph{Zwitscherkasten}, a low-cost, multimodal system for avian species monitoring that combines audio- and image-based deep learning pipelines on resource-constrained edge devices. By leveraging lightweight architectures, event-driven inference, and local deployment on Raspberry~Pi–class hardware, we demonstrate that accurate and non-invasive bird monitoring is feasible without reliance on cloud infrastructure. The proposed system supports continuous operation, real-time user feedback, and do-it-yourself deployment, making it suitable for citizen science and home-based biodiversity monitoring.

Experimental results across multiple model families highlight a fundamental trade-off between classification accuracy and computational efficiency on embedded platforms. While state-of-the-art architectures such as PaSST and medium-scale YOLO variants achieve strong performance, sustained real-time inference-particularly for continuous visual processing-remains challenging under strict CPU and memory constraints. These observations motivate architectural and optimization choices that prioritize low-duty-cycle, event-triggered inference over continuous streaming.

Future work will focus on improving deployment efficiency through systematic model compression techniques, including post-training and quantization-aware training, structured and unstructured pruning, and knowledge distillation from larger teacher models. In particular, low-bit integer quantization and sparsity-aware execution offer promising avenues for reducing latency and energy consumption while preserving accuracy. Additionally, further optimization of the visual pipeline-such as adaptive frame sampling, dynamic resolution scaling, and hardware-specific acceleration-could enable more robust real-time performance on edge devices.

Beyond performance optimization, future extensions include long-term field evaluations, improved sensor fusion between audio and visual modalities, and support for additional taxa. Together, these directions aim to advance practical, scalable, and accessible edge-based biodiversity monitoring systems.


\section*{Acknowledgment}

The authors acknowledge the use of AI assisted tools including ChatGPT, Gemini, GitHub Copilot and Perplexity for language editing, code suggestions, and supplementary research insights.





\printbibliography

\end{document}